\documentclass[conference]{IEEEtran}
\IEEEoverridecommandlockouts
\usepackage{cite}
\usepackage{amsmath,amssymb,amsfonts}
\usepackage{algorithmic}
\usepackage{graphicx}
\usepackage{textcomp}
\usepackage{xcolor}
\usepackage{blindtext}
\usepackage{todonotes}

\usepackage{siunitx}
\usepackage{acronym}
\usepackage{listings}
\usepackage{graphicx}
\usepackage{wrapfig}
\usepackage{booktabs}
\usepackage{caption}
\usepackage{subcaption}
\usepackage[hidelinks]{hyperref}

\usepackage{tikz}
\usetikzlibrary{fit,positioning,arrows,automata,shapes,shadows,fadings, decorations.pathreplacing}
\usepackage{tikz-network}

\newcommand{\vicon}[0]{Vicon~}
\newcommand{\sensorf}[0]{\emph{Sensor Floor~}}
\newcommand{\wrt}{w.r.t.\@\xspace}

\newcommand{\sensors}{\ensuremath{\mathcal{S}}}
\newcommand{\nodes}{\ensuremath{\mathcal{N}}}
\newcommand{\buffer}{B}
\newcommand{\sensordata}{\ensuremath{\mathcal{B}}}
\newcommand{\vicondata}{\ensuremath{\mathcal{V}}}
\newcommand{\framedata}{\ensuremath{\mathcal{F}}}
\newcommand{\data}{\ensuremath{\mathcal{D}}}
\newcommand{\x}{\ensuremath{\vec{x}}}

\newcommand{\y}{\ensuremath{\vec{y}}}
\newcommand{\ts}{t}

\newacro{rssi}[RSSI]{Received Signal Strength Indicator}    
\newcommand{\rssi}{\ac{rssi}~}
\newacro{acel}[Accel]{Acceleration}    

\newacro{av}[AV]{Angular Velocity}    

\newacro{mf}[MF]{Magnetic Field Strength}    

\newacro{rtt}[RTT]{Round Trip Time}    
\newcommand{\rtt}{\ac{rtt}~}
\newacro{CNN}[CNN]{Convolutional Neural Network}    

\newacro{rpi}[RPi]{Raspberry PI}    
\newcommand{\rpi}{\ac{rpi}}

\acrodef{WSN}[WSN]{Wireless Sensor Network}    
\newcommand{\WSN}{\ac{WSN}~}

\acrodef{IoT}{Internet of Things}    
\newcommand{\IoT}{\ac*{IoT} }

\acrodef{6LoWPAN}[6LoWPAN]{IPv6 over Low-Power Wireless Personal Area Networks}

\acrodef{SoC}[SoC]{System on Chip}
\newcommand{\soc}{\ac{SoC}\xspace}

\acrodef{MCU}[MCU]{Micro-Controller Unit}
\newcommand{\mcu}{\ac{MCU}\xspace}

\acrodef{IMU}[IMU]{Inertial Measurement Unit}
\newcommand{\imu}{\ac{IMU}\xspace}

\acrodef{CPS}[CPS]{Cyber-Physical Systems}
\newcommand{\CPS}{\ac{CPS}\xspace}

\acrodef{MQTT}[MQTT]{Message Queuing Telemetry Transport}
\newcommand{\MQTT}{\ac{MQTT}\xspace}

\xdefinecolor{tugreen}{RGB}{132, 184, 24} 
\xdefinecolor{tudarkgreen}{RGB}{100, 150, 0} 
\xdefinecolor{tulightgreen}{RGB}{224, 234, 204} 
\xdefinecolor{tugray}{RGB}{207, 208, 210} 
\xdefinecolor{tugraybg}{RGB}{178, 179, 204} 
\xdefinecolor{tuorange}{RGB}{227, 105, 19} 
\xdefinecolor{tuyellow}{RGB}{242, 189, 0} 
\xdefinecolor{tucitron}{RGB}{249, 219, 0} 
\xdefinecolor{tublue}{RGB}{46, 134, 171} 
\xdefinecolor{tudarkblue}{RGB}{30, 89, 114} 
\xdefinecolor{tuviolet}{RGB}{61, 52, 139} 
\xdefinecolor{tumagenta}{RGB}{139, 52, 88} 
\xdefinecolor{tudarkergreen}{RGB}{75, 98, 44} 
\xdefinecolor{tuolive}{RGB}{83, 145, 45} 
\xdefinecolor{tulime}{RGB}{215, 215, 0} 
\xdefinecolor{tugraygreen}{RGB}{217,233,229} 

\tikzfading[name=arrowfading, top color=transparent!0, bottom color=transparent!95]
\tikzset{arrowfill/.style={top color=OrangeRed!20, bottom color=Red, general shadow={fill=black, shadow yshift=-0.8ex, path fading=arrowfading}}}
\tikzset{arrowstyle/.style={draw=FireBrick,arrowfill, single arrow,minimum height=#1, single arrow,
single arrow head extend=.4cm,}}
\tikzstyle{block} = [rectangle, draw, fill=tugreen!20, text centered, rounded corners, minimum height=4em, minimum width=7cm]
\tikzstyle{oblock} = [rectangle, draw, fill=tuorange!20, text centered, rounded corners, minimum height=4em, minimum width=7cm]
\tikzstyle{sblock} = [rectangle, draw, fill=tuorange!20, text centered, rounded corners, minimum height=4em, minimum width=2cm]
\tikzstyle{line} = [draw, thick, -latex']

\makeatletter
\newcommand\inlinecomment{\begingroup\@makeother\#\@inlinecomment}
\newcommand\@inlinecomment[2]{%
    \todo[%
        inline,%
        color=#1,%
    ]{#1:  #2}%
    \endgroup%
}%
\newcommand\comment{\begingroup\@makeother\#\@comment}
\newcommand\@comment[2]{%
    \todo[%
        color=#1,%
    ]{#1:  #2}%
    \endgroup%
}%
\makeatother

\colorlet{DH}{blue!10}    

\usepackage{comment}
\def\BibTeX{{\rm B\kern-.05em{\sc i\kern-.025em b}\kern-.08em
    T\kern-.1667em\lower.7ex\hbox{E}\kern-.125emX}}
    
\def\thesubsubsectiondis{\unskip\arabic{subsubsection}).}

\begin{document}

\title{A Grid-based Sensor Floor Platform for Robot Localization using Machine Learning
}
\author{
    \IEEEauthorblockN{Anas Gouda$^{1}$, Danny Heinrich$^{2}$, Mirco Hünnefeld$^{3}$, Irfan Fachrudin Priyanta$^{1}$, Christopher Reining$^{1}$, Moritz Roidl$^{1}$}
    \IEEEauthorblockA{$^{1}$Chair of Material Handling and Warehousing, TU Dortmund}
    \IEEEauthorblockA{$^{2}$Chair of Artificial Intelligence, TU Dortmund}
    \IEEEauthorblockA{$^{3}$Astroparticle Physics Group, TU Dortmund}
    \ firstname.lastname@tu-dortmund.de
}

\maketitle

\begin{abstract}
\WSN applications reshape the trend of warehouse monitoring systems allowing them to track and locate massive numbers of logistic entities in real-time. 
To support the tasks, classic Radio Frequency (RF)-based localization approaches (e.g. triangulation and trilateration) confront challenges due to multi-path fading and signal loss in noisy warehouse environment. 
In this paper, we investigate machine learning methods using a new grid-based \WSN platform called \sensorf that can overcome the issues. 
\sensorf consists of 345 nodes installed across the floor of our logistic research hall with dual-band RF and \imu sensors. Our goal is to localize all logistic entities, for this study we use a mobile robot. We record distributed sensing measurements of \rssi and \imu values as the dataset and position tracking from Vicon system as the ground truth. The asynchronous collected data is pre-processed and trained using Random Forest and Convolutional Neural Network (CNN). The CNN model with regularization outperforms the Random Forest in terms of localization accuracy with $\approx \SI{15}{cm}$. Moreover, the CNN architecture can be configured flexibly depending on the scenario in the warehouse. 
The hardware, software and  the CNN architecture of the \sensorf are open-source under \url{https://github.com/FLW-TUDO/sensorfloor}.
\end{abstract}

\begin{IEEEkeywords}
distributed sensing, machine learning, wireless sensor networks, localization, resource constraint
\end{IEEEkeywords}

\section{Introduction}


Nowadays, \CPS and \IoT devices are widely being used to digitize industrial processes in the logistic sector~\cite{falkenberg2017phynetlab}. Typical logistic warehouse operation is identical to a collaborative space between humans and swarm robots that perform sorting, picking, or transporting parcels. A \WSN based localization is required to monitor massive numbers of logistic entities in real-time and thus enabling seamless warehousing processes\cite{jiang_logistics_2020}. To accomplish those goals, a precise indoor localization system with \si[]{\centi\metre} accuracy range must be investigated. 

RF-based indoor localization system suffers from getting reliable position and are prone to signal loss in the multi-path industrial environment, especially for classical methods such as triangulation, trilateration, and statistical model \cite{paulavicius_temporal_2022, ayyalasomayajula_deep_2020}. In addition, a large number of asynchronous data sources in \WSN appears to be a challenge\cite{redhu_joint_2018}. Thus, it is also critical to provide strategies and concepts for distributed systems to address both problems.

Machine learning has the capability to handle indoor localization tasks \cite{ye_wireless_2020}. In the case of warehouse environments, machine learning also has the potential to manage the high noise levels. Authors in \cite{lee2012separable} introduced distributed sensing methods for classification or regression tasks. In this work, we propose machine learning based indoor localization methods that can model such noisy environments and locate objects accurately.
We develop the application using a new distributed grid, asynchronous data source which is called the \sensorf platform. By doing so, we are able to conduct the experiment based on industrial practice where the whole area of the  warehouse is covered by the \sensorf nodes.
 Finally, the integration of distributed sensing concept and machine learning methods will be discussed furthermore in order to estimate global position of mobile robots.

The rest of this paper is organized as follows: Section \ref{overview} presents the related works of \sensorf and RF-based localization. Section \ref{sensorfloor} describes the \sensorf architecture, hardware, and software stack. Afterward, section \ref{experiment_results} specifies the experiment setups of data acquisition, training the \sensorf dataset, and the results of localization accuracy. Finally, section \ref{conclussion} summarizes the current work and future outlook.





\section{Related Works} \label{overview}
This section briefly introduces the related works for the development of \sensorf as well the RF-based localization.
The two following examples of Sensor Floor are examples of floor-augmented sensor applications called ”magic carpet” that used pressure-sensitive areas for measuring the interactions on the floor \cite{paradiso2000sensor, richardson2004z}. The first version of "magic carpet" \cite{paradiso2000sensor} used woven piezo-electric wires while the next version \cite{richardson2004z} has been improved in terms of the addressable nodes so that they can be deployed in any shape or size by utilizing an array of force-sensitive resistors on each node.

Classical RF-based localization system falls into several techniques, for instance, direct methods and RSSI-based. 
Experiments in \cite{tang_comparison_2019} compare diverse indoor positioning approaches of Channel State Information (CSI) and RSSI-based: fingerprinting, trilateration, sequence-based localization (SBL). The results reveal that RSSI-based methods prone to signal noises and exhibit root-mean-square error (RMSE) of position between $\SI{0.8}{-}{3.3}{m}$ whilst CSI has a position accuracy of $\approx \SI{0.3}{m}$.
Direct method i.e., Ultra-Wideband (UWB) position estimation based on Time Difference of Arrival (TDOA) demonstrates limitations in the case of a Non-Line-of-Sight (NLOS) environment with RMSE of $\SI{67}{\%}$ as shown in \cite{garcia_robust_2015}.

To improve the accuracy of the indoor localization system, \cite{ayyalasomayajula_deep_2020, paulavicius_temporal_2022} initiate to integrate RSSI-based localization with machine learning methods. 
A person localization system using RSSI along with self-supervised learning is developed by \cite{paulavicius_temporal_2022} for the application in residential homes. Bluetooth Low Energy (BLE) device is equipped by a person to transmit beacon signals to the distributed APs at home. Time-aligned BLE RSSI values and annotated positions were trained using the k-Nearest-Neighbour (kNN) machine learning model resulting in an F1 score of $ \SI{0.89}{\%}$ at room level. Deep Neural Network (DNN) and Wi-Fi measurements are integrated into a system named DLoc \cite{ayyalasomayajula_deep_2020}. The encoder-decoder architecture of DLoc is capable to generalize the environment space using the heatmap image. DLoc accomplishes $\SI{0.7}{cm}$ localization error in  a space of 500 sq. ft. To summarize, appropriate selections of architecture in machine learning model must be considered to achieve precise location in the \si[]{\centi\metre} range.

\section{Sensor Floor}
\label{sensorfloor}

\begin{figure}[htb!]
    \centering
    \includegraphics[width=0.47\textwidth]{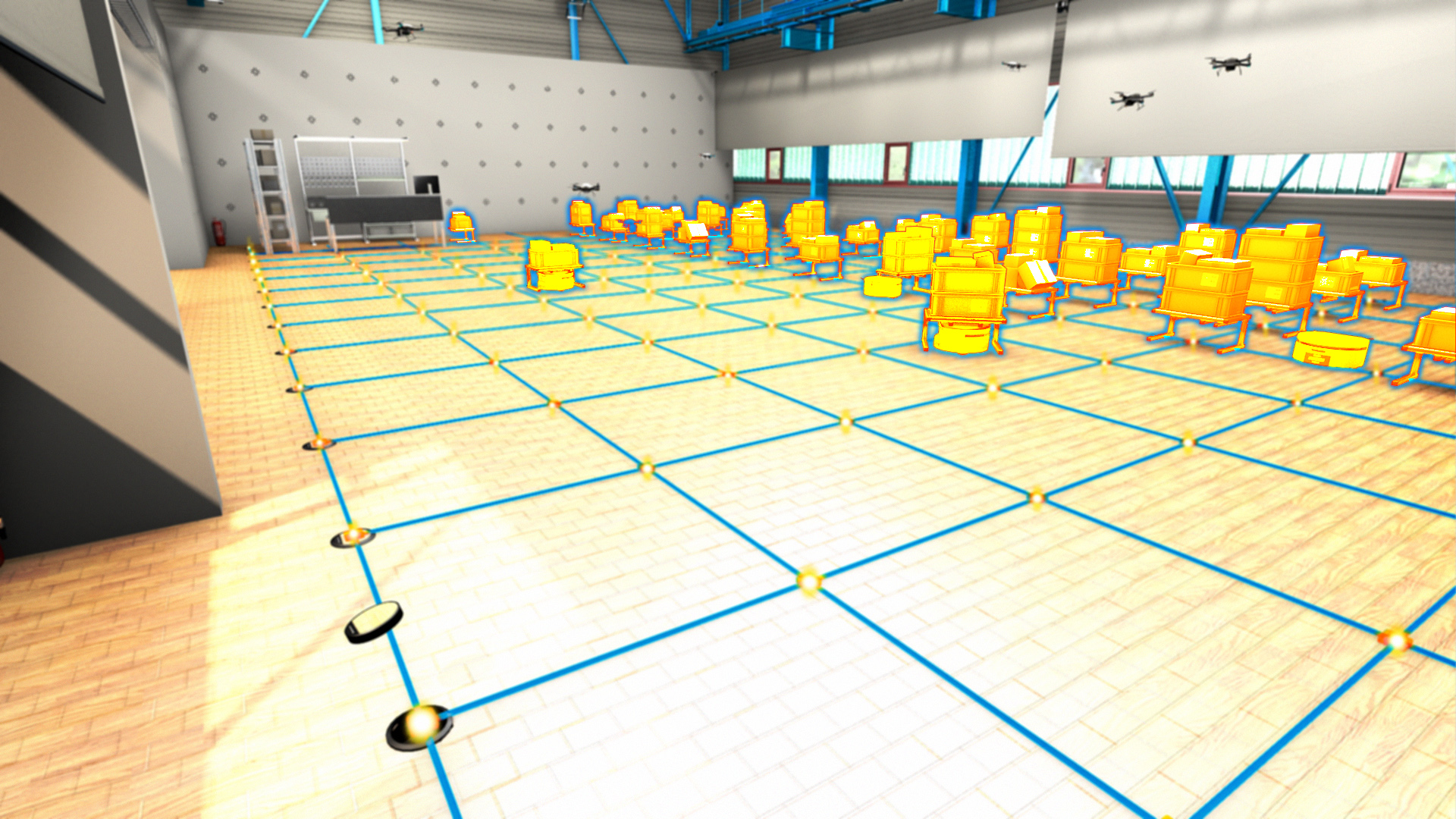}
    \caption{Illustration for the placement of nodes of the \sensorf}
    \label{fig:sensor_floor_render}
\end{figure}
\vspace{-1em}

\sensorf is a low-power distributed sensing platform that can be used for the development and evaluation of different applications. In our previous work, we evaluated our radio flooding protocol in \cite{10.1007/978-3-030-71061-3_6} using the \sensorf. In this work, we use it to develop our machine learning-based localization methods. The Sensor Floor has three components:
the CC1350 SensorTag, a custom-made breakout board where the communication bus and power lines are connected, and a sink computer for data acquisition and firmware flashing. The \sensorf is applied as a grid of 1x1\SI {}{\metre\squared} cells that are deployed in a 30x15\SI{}{\metre\squared} (length X width) hall, with 23 strips spanning the \SI{30}{\metre} length and each strip has 15 nodes covering the whole \SI{15}{\metre} width. Each strip is composed of 15 sensor nodes connected with 12 volts power supply, 1-wire and RS422 communication lines. One end of the 15-meter strip terminates with two USB converter for a 1-wire bus and the RS422 bus to be connected to the sink computers. In total, there are 345 nodes in 23 strips, and 23 internet-connected \acp{rpi} as the sink computer. Each node can be flashed individually, and the whole floor can be flashed in parallel using a command-line tool developed exclusively for this deployment. The \acp{rpi} sink computers are time-synchronized and connected to power and network using a PoE connection. The data acquisition is tested to be synchronized within a few milliseconds within the allowable tolerance for data acquisition. The nodes' synchronization messages are delivered every 4 seconds on an average due to the design choices of using a 1-wire bus to apply changes to the communication bus.


\begin{figure}[htb!]
    \centering
    \vspace{-0.2cm}
    \includegraphics[width=0.45\textwidth]{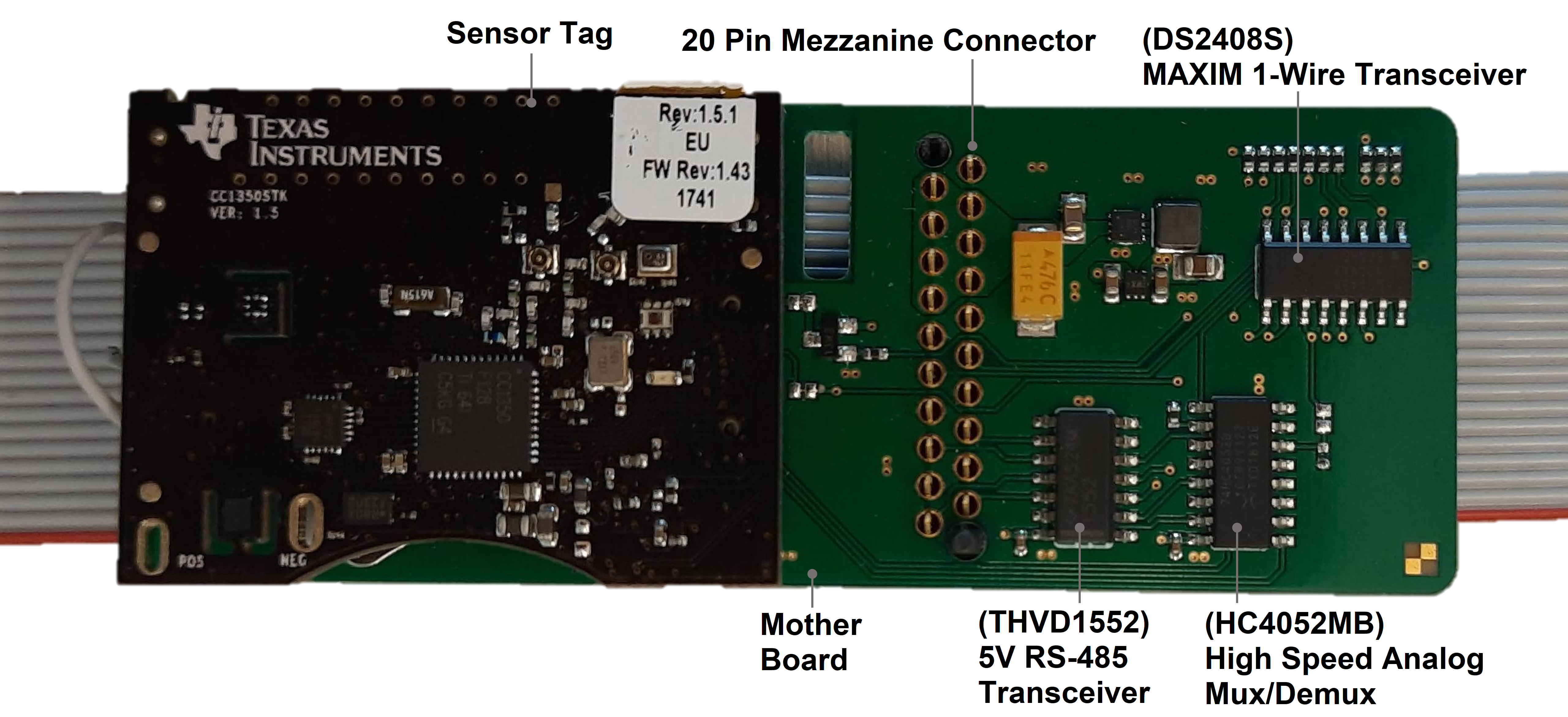}
    \caption{Each \sensorf node consists of a CC1350 SensorTag (black kit) and a custom made breakout board (green kit)}
    \vspace{-0.4cm}
    \label{fig:stk_and_breakout}
\end{figure}

A single node is chosen which has ultra-low-power \WSN specifications. Each node is comprised of two boards, the breakout board, and a Texas Instruments CC1350 SensorTag.
This device is an off-the-shelf \IoT hardware product that was developed for prototyping purposes. The breakout board is used for deployment and supplies the SensorTag with power, data, and flashing connections. CC1350 SensorTag features the CC1350 \soc as the \mcu and is equipped with 10 low-power MEMS sensors and a Dual-band radio. The radio supports both Sub-1GHz band and 2.4GHz band. CC1350 SensorTag has 10 on-board sensors, but in this work, we use only IMU sensors (Accelerometer, Gyroscope, Magnetometer).
There are two types of software for the stack to develop applications.
First is the management tool containing boilerplate code. The second type is the application-specific firmware developed for the CC1350 hardware, with guidelines for flashing the nodes.
The management software includes open source tools like the SBL flasher, which abstracts flashing for a CC1350 node using the serial boot loader interface.
The software is distributed so that all of the sink computers can run the same software, and their IP addresses are used to run commands remotely.
A frequently run command is to synchronize the time between the sink computers before starting an experiment to reduce the skew in synchronization.


The 1-wire bus status is exposed using the \emph{owfs} software package, which abstracts the 1-wire communication into a set of file system changes.
The \emph{imu\_reader} is a tool that contains a hex file targeted for the CC1350 platform. It reads the IMU data and the RSSI values of any received messages and transmits when an interrupt is received.
This software also implements the boilerplate code for developing full-stack \sensorf applications. The 1-wire nodes are queried in a loop every 4 seconds, accounting for the 1-wire communication delays.
The 1-wire communication involves turning off the previous node that communicated using the RS422 bus and turned on the next device in the strip to communicate.
Once the sink computer receives the messages, it is directly posted into multiple \MQTT topics built with various unique identifiers, as shown in followings: \emph{/imu\_reader/(MAC address sink)/(Node ID)} and \emph{/imu\_reader/(strip id)/(Node ID)}.

This allows for platform-independent, language-agnostic distributed software architecture for developing \sensorf applications.
A simple GET request responds with a JSON formatted string that hosts all the payload information of each node and each strip and is available on-demand over the internet.

\section{Experiment and Results} \label{experiment_results}
\subsection{Experiment: Data Acquisition}

\begin{figure}[!htb]
    \centering
    \includegraphics[width=0.65\linewidth]{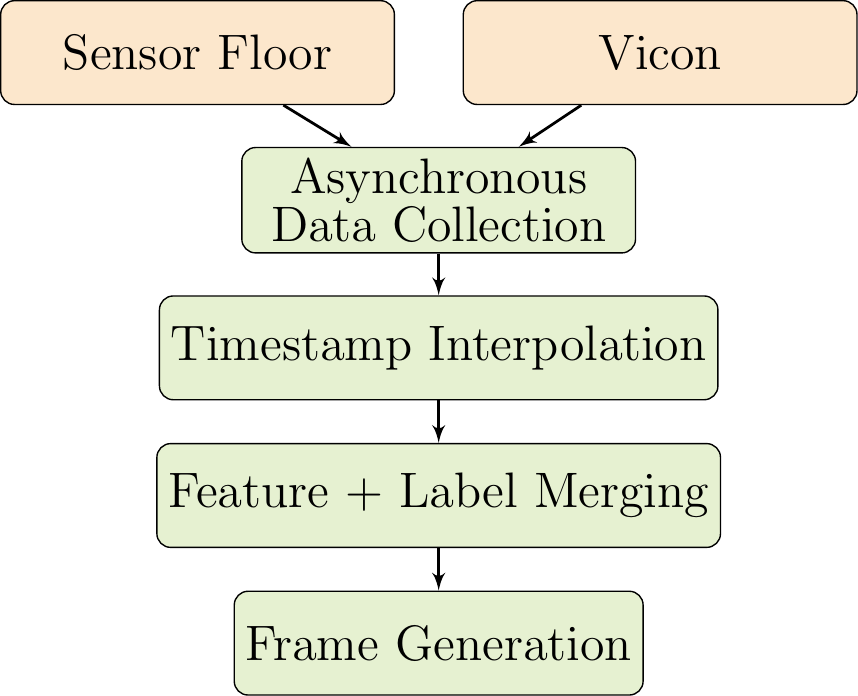}
    \caption{Preprocessing pipeline of \sensorf and \vicon data.}
    \vspace{-0.4cm}
    \label{fig:pipeline}
\end{figure}

Data collection and processing are vital aspects of the full hardware and software stack.
The data collection and preprocessing step of spatio-temporal data are divided into several steps and illustrated in figure \ref{fig:pipeline}.
The technical details about both systems used to collect the data are given in table \ref{tab:system_specs}. The pipeline consists of three elements, taking  data generated by \sensorf, and \vicon as input and outputs a set of processed frames, containing data from every sensor, a label and a timestamp. \vicon marker-based motion capturing systems from is used to provide a reference system to track and locate the marked entities with a \si[]{\milli\metre} precise location.


\subsubsection{Data Collection}
We collected the data during a single day, using a mobile robot with a speed of approximately \SI{1}{\meter\per\second}, equipped with a magnet and signal emitter.
The data is divided into nine separate training runs consisting of vertical, horizontal, and diagonal movement patterns.
By covering the entire accessible \sensorf area it was ensured that all sensors located inside the arena were able to record data with the robot being in the node's vicinity.
The data used for evaluation purposes was recorded separately to ensure a strict separation of training and test data.
By introducing randomized movement patterns for the test data, a high correlation between training and evaluation data is avoided, allowing for an unbiased evaluation of our models.

\begin{table}[!htb]
    \caption{Sample rates, number of data sources and measurements collected for both \vicon and \sensorf.}
    \centering
    \resizebox{\columnwidth}{!}{%
    \begin{tabular}{c||c|c}
    \toprule
     & \textbf{SensorFloor} & \textbf{Vicon} \\
    \midrule
    Sample Rate & \SI{6}{\hertz} & \SI{200}{\hertz} \\
    \midrule
    Data Sources &{$\!\begin{aligned} 
               \sensors = \{1, \ldots, 23\}, &\nodes = \{1, \ldots, 15\}, \\
               \sensors \times \nodes &= \mathcal{P},  \\
               |\mathcal{P}| = 345 \;&\text{Nodes} \end{aligned}
   $}& 1 Tracking System \\
    \midrule
            & Accelerometer (x,y,z) ~[\si[]{\metre\per\second\squared}] &  \\
    Data Recorded    & Gyroscope (x,y,z) ~[\si[]{\degree\per\second}] &Position (x,y,z)  ~[\si[]{\milli\meter}]  \\
            & Magnetometer (x,y,z) [\si[]{\micro\tesla}] &  Rotation (x,y,z) [\si[]{\radian}]  \\
            & RSSI ~[\si[]{\deci\bel}m] &   \\
    \midrule
    Data Set & 
     {$\!\begin{aligned} 
               \mathcal{B}_{s,n} &= \{(B^0, t^0), \ldots, (B^l, t^l)\}  \\
               B^i &= (\x^0, \ldots, \x^{k_i}) \end{aligned}$}
        & $\mathcal{V} = \{(\y^1, \ts^1), \ldots, (\y^n, \ts^n)\}$ \\
    \midrule
        \# Features/Labels & 3450 & 6 \\
    \bottomrule
    \end{tabular}
    }
    \label{tab:system_specs}
\end{table}

\begin{figure}
\centering

\begin{subfigure}[htb]{0.50\textwidth}
    \centering
    \includegraphics[width=\linewidth]{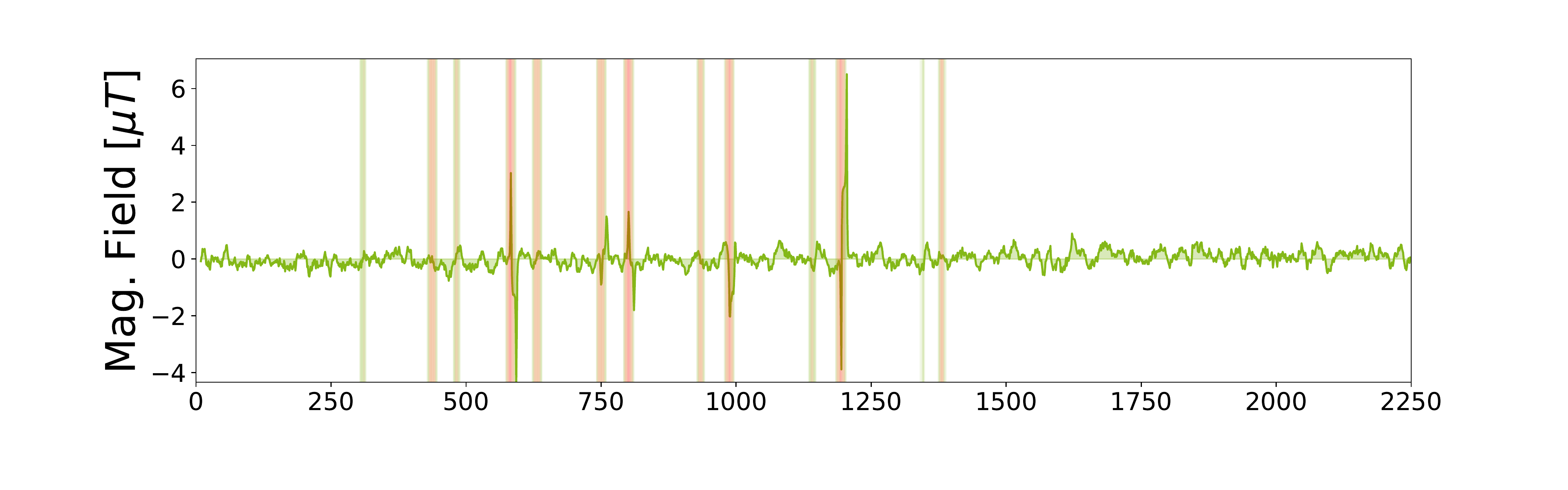}
\end{subfigure}

\vspace{-1em}%
\begin{subfigure}[htb]{0.50\textwidth}
    \centering
    \includegraphics[width=\linewidth]{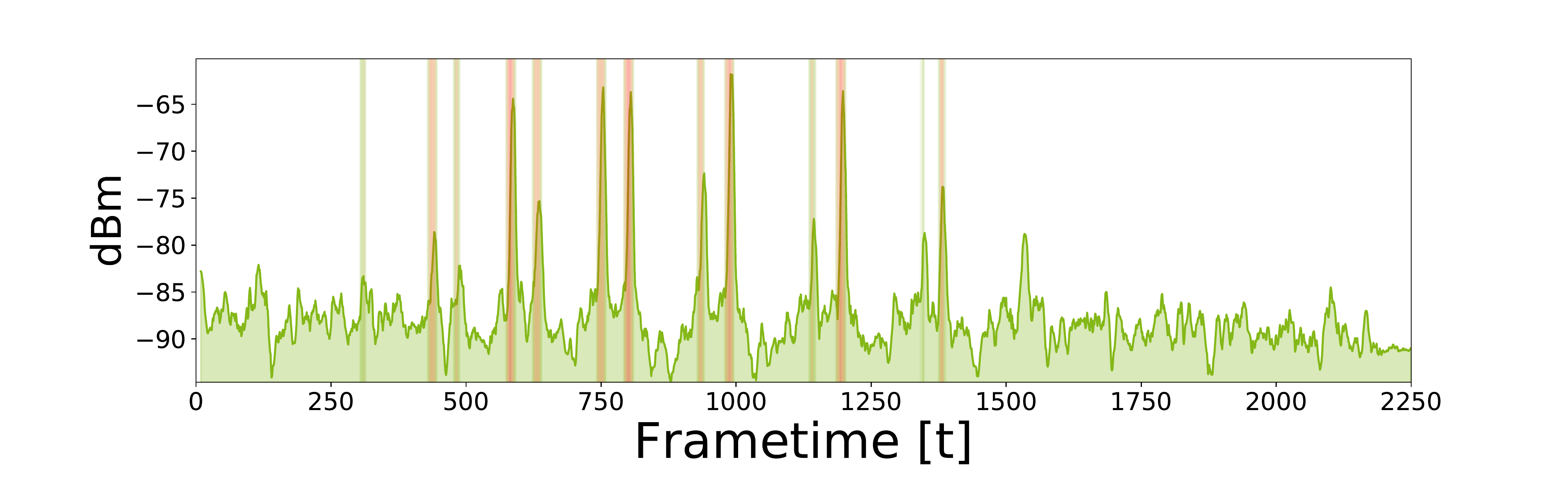}
\end{subfigure}
\caption{Visualization for \rssi measurement and Magnetometer measurement in x-direction.}
\vspace{-0.6cm}
\label{fig:timeseries}
\end{figure}

Recall that each node periodically generates a set of ten measurements, which are written sequentially to a local buffer. 
The data is queried in a round-robin style by each \rpi~ with a \rtt of \SI{4}{\second}, flushing the buffer and creating a payload bundle with measurements and a single timestamp.
Examples of sensor measurements for a single node are shown in figure \ref{fig:timeseries} for the \acp{IMU} unit and \rssi value.
The distance of the robot to the node is indicated by the vertical lines changing from yellow to red when the robot moves closer to the node [$\text{Red} \leq \SI{1}{\metre}, \; \text{Yellow} \leq \SI{2}{\metre} \leq \text{Green} \leq \SI{3}{\metre}$]. As a result, \rssi and magnetometer measurements are increased when the robot is nearby.



Initially, features and labels are obtained separately.
Features are obtained from \sensorf and the labels are obtained from the \vicon system.
By using the \vicon tracking system, labels are automatically generated and only have to be matched to the \sensorf observations based on the timestamps generated by both systems.
However, due to the asynchronous nature of both systems, additional preprocessing steps are necessary to create synchronized training and test sets.


By assigning additional identifiers to messages from each node, they can directly be mapped to their source, keeping the spatio-temporal information.
Now, let 
\begin{equation}
\mathcal{P} = \sensors \times \nodes = \{(1,1), (1,2), \ldots (23,15)\}
\end{equation}
be the set of nodes as defined in table \ref{tab:system_specs}.
Recall that every node asynchronously records and writes the measurements to its local buffer.
While their system clocks are synchronized (c.f. section \ref{sensorfloor}), the buffers are read and written asynchronously, assigning a timestamp to every incoming data.
Additional preprocessing steps are required to synchronize and match the \sensorf measurements relative to the label.

\subsubsection{Timestamp Interpolation}
First, timestamps of each individual element are generated by linearly interpolating between existing timestamps.
Batches may contain a different number of measurements as the buffer may be read at slightly varying times due to real-world conditions.
New timestamps are then interpolated between the previous and the current timestamp using an equidistant step size.
More precisely, let 
\vspace{-0.4em}
\begin{equation}
    \sensordata_{s,n} = \{(\buffer^1, \ts^1), (\buffer^2, \ts^2), (\buffer^k, \ts^{k})\}, \quad (s,n) \in \mathcal{P}
\end{equation} 
be a set of size $k$ with measurements $\buffer^i = \{\x^1, \x^2, \ldots, \x^{l_i}\}$ and timestamps $\ts$ with $\ts_{i-1} < \ts_{i}, \; 1 \leq i \leq k$ obtained from an arbitrary node.
Missing timestamps are generated by linearly interpolating between $\ts^i$ and $\ts^{i-1}$, based on the measurements observed during the time between two buffer reads.
Timestamps are interpolated using equidistant time steps between the previous and current buffer time:
\begin{equation}
    \Delta_{\ts^i} = (\ts^i - \ts^{i-1}) / \lvert \buffer^{i} \rvert.
\end{equation}
The interpolated timestamp for the $j$th element in a buffer is obtained by $\hat{\ts}^{i}_j = \ts^{i-1} + j \cdot \Delta_{\ts^i}$.
Generating the timestamps for the first buffer element the standard \rtt of \SI{4}{\second} was chosen.
By interpolating the timestamps for every measurement on every node, labels and measurements, outputs from \vicon and \sensorf are matched.
Since \vicon generates timestamps for all measurements, the interpolation is only applied to the \sensorf data. The buffers from each node are flattened, obtaining a multivariate time series with ten elements:
\begin{equation}
    \begin{split}
    \hat{\sensordata}_{s,n} &= \{(\x^0, \ts^0), (\x^1, \ts^1), \ldots,  (\x^l, \ts^l)\}, \\
    \end{split}
\end{equation}
where each node contains $l_{s,n}$ elements.
Note that the number of elements differs between nodes due to the real-world conditions mentioned previously. 

\subsubsection{Merging \sensorf and \vicon}
Using the augmented data containing interpolated timestamps, the batches $\hat{\sensordata}_{s,n}$ are used to combine \vicon and \sensorf data.
Recall that by definition from table \ref{tab:system_specs} the \vicon data $\vicondata$ contains label and time information.
Samples from \sensorf and \vicon are matched based on the closest \vicon timestamp \wrt the interpolated \sensorf timestamps.
More formally for every node $s,n \in \mathcal{P}$, a label $y^{j'} \in \vicondata$ is assigned to each measurement $(\x^i, \ts^i) \in \hat{\sensordata}_{s,n}$ such that:
\vspace{-0.4em}
\begin{gather}
    \y^i_{s,n} = \y^{j'}, \\
    j' = \arg\min_{j} \;\lvert \ts^i_{s,n} - \ts^j \rvert , \quad (\y^j,\ts^j) \in \vicondata
\end{gather}
the timestamp from the \vicon data set is closest to the time $\ts^i$.
The data for each node, containing features, timestamps and labels is then obtained as:
\vspace{-0.4em}
\begin{equation}
    \label{eq:merge-sensor-vicon}
    \data_{s,n} = \{(\x^0,\y^0, \ts^0), (\x^1,\y^1, \ts^1), \ldots, (\x^l,\y^l, \ts^l)\}.
\end{equation}


\subsubsection{Frame Generation}

Timestamp-augmented labelled data sets are merged into a set of frames.
By choosing the node with the least number of observations $l_{\min} = \min_{s,n} |\data|_{s,n}, \; \forall s,n \in \mathcal{P}$ as reference point, creating multiple frames with duplicate measurements is avoided.
For each reference point and based on the reference timestamp one measurement from each of the remaining nodes is matched.
Let $\framedata$ denote the sequence of frames $F$, with measurements $\x^{j_{s,n}}$ from each node some time $\ts$:
\vspace{-0.4em}
\begin{gather}
    \framedata = \{F^1, F^2, \ldots, F^{l_{min}}\} \\
    F^i = \{\data_{1,1}^{j}, \data_{1,2}^{j}, \ldots, \data_{23,15}^{j}\} \\
    j_{s,n} = \arg\min_{j} \quad \lvert t^j - \bar{t}^i \rvert \quad \forall \;  0 \leq i < l_{\min},
\end{gather}
where $j_{s,n}$ is the index used to select the features from $\data_{s,n}$.
Briefly, measurements from other nodes are merged based on the timestamp $t^j$ closest to the timestamp $\bar{t}^i$ obtained from the data set with the least amount of observations.
Note that each frame $F^i$ now contains multiple labels and timestamps, which are averaged into a single label and timestamp for the machine learning task. 
The full spatio-temporal data set is then given by the collection of features from every node, an averaged label and timestamp:
\begin{equation}
\begin{aligned}
    \data = \{(\{\x_{s,n}^0 | s,n \in \mathcal{P}\}, \bar{\y}^0, \bar{\ts}^0), \ldots, \\
    (\{\x^{l_{min}}_{s,n} | s,n \in \mathcal{P}\}, \bar{y}^{l_{min}}, \bar{\ts}^{l_{min}})
    \}.
\end{aligned}
\end{equation}
Due to technical reasons, some nodes experience delay when first starting the measurements.
Leading and trailing observations are cut off to create matching start and end points.

\begin{figure}
\includegraphics[width=\linewidth, trim=0 0 0 0, clip]{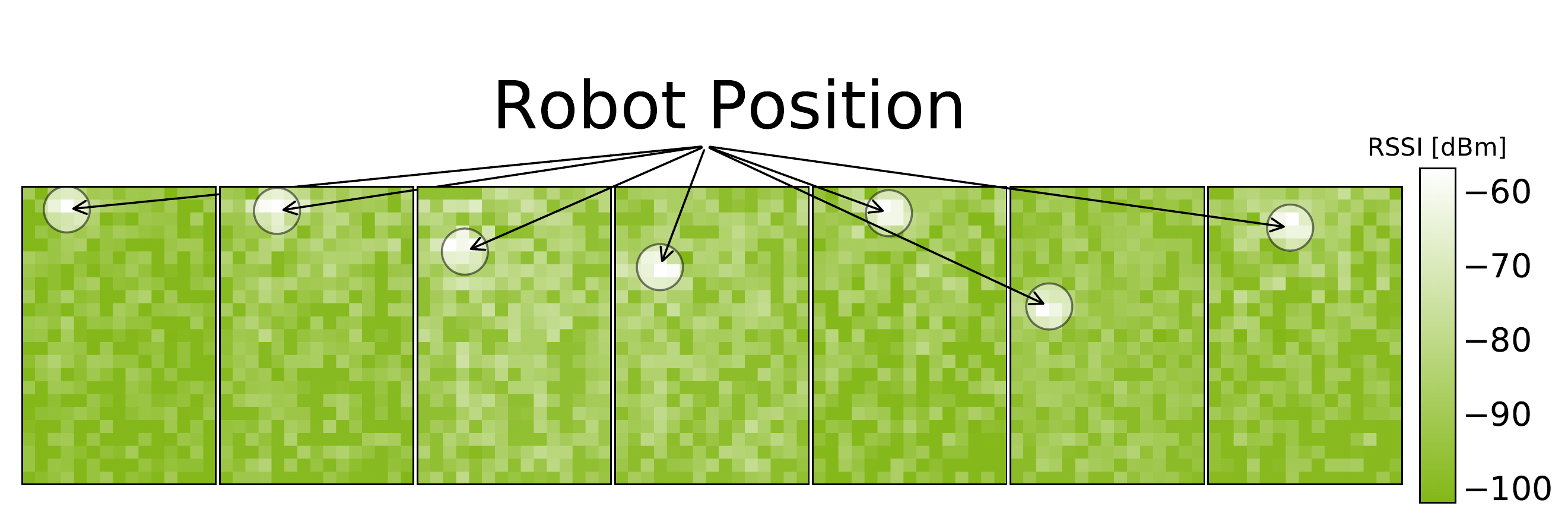}\\
\caption{Bird's eye view of the \sensorf area recreated from processed sensor data. Here the \rssi data is shown for several generated frames with different robot positions also shown. Increasing \rssi values are shown from green to white, with more white colors corresponding to a higher signal strength}
\vspace{-0.5cm}
\label{fig:heatmap}%
\end{figure}

The generated frames are visualized in figure \ref{fig:heatmap}, where 15 consecutive frames with \rssi feature and the robot position are shown. Each frame represents the measurements from all nodes for a fixed time $\ts$, with a total of \texttt{3450} features and \texttt{2} labels.
Labels, other than x and y-coordinates, i.e., rotation and z-axis were dropped in this context.

\subsection{Experiment: Training Sensor Floor Dataset for Localization}

For the localization task, we combine position tracking and distributed sensing, defining a regression task on the two-dimensional position of a moving object (robot).
Recall that $\x_{s,n}^i$ are ten features observed on node s,n at some time $\ts^i$.
The localization task is approach by employing two different machine learning methods: a traditional random forest approach and a \ac{CNN}.
The CNN is later extended with an additional preprocessing step adding a regularizing penalty term on velocity and acceleration of the predicted positions.

\subsubsection{Localization via Random Forest}


Random Forests are ensemble methods, which use a collection of Decision Trees.
Decision trees generate rules from features forming a binary tree, which is used for classifying new samples.
For more in-depth information about this type of machine learning model we refer the interested reader to Friedman et al. \cite{friedman2001elements}.
These supervised machine learning models have been used for both classification and regression tasks  and are known to perform well on different data sets\cite{fernandez2014we}.

The training and test data sets for the random forest are additionally preprocessed using a scaling method and a feature aggregation method.
First, the data is scaled to the range of $[0,1]$ using a min-max scaler applying the following transformation:
\begin{equation}
    \x'_i = \frac{\x_i - \min(\mathcal{D}_{\cdot i})}{\max(\mathcal{D}_{\cdot i}) - \min(\mathcal{D}_{\cdot i})},
\end{equation}
where $\x_i$ is a single feature on each node.
Based on the normalization, new features are created using not only a single node, but also its direct neighbors.
Recall, that the sensors are arranged in a grid, i.e., each node has at most eight direct neighbors.
By summing over each feature of the node neighborhood a broader view on the sensor grid is created, incorporating more spatial information into each feature.
Note that the feature extraction applied does not consider weights for each sensor, i.e., all observations are weighted equally:
\begin{equation}
    \vec{\tilde{x}}_{s,n} = \x'_{s,n} + \sum_{\vec{v}' \in \mathcal{N}(\x'_{s,n})} \vec{v}' \quad \forall (s,n) \in \mathcal{P},
\end{equation}
where $\mathcal{N}$ is the neighborhood function and $\x'_{s,n}$ are ten scaled features from each node. 
The scaled and transformed features are then used as input to the random forest model.

Random forests are trained with the python machine learning library Scikit-Learn \cite{scikit-learn} using a 10-fold cross-validation split and a hyper-parameter search for optimal number of trees and tree depth.
Acceleration and angular velocity were dropped following an initial set of experiments as both features did not contribute significantly to feature importance.
Hence, only magnetometer and \rssi features were used for training. 



\subsubsection{Localization via Convolutional Neural Networks}
\label{ssec:CNN}
In addition to the traditional machine learning approach via a Random Forest, 
a deep learning model is employed for the regression task.
The \sensorf consists of a regular grid of $23 \cdot 15$ sensors and is therefore well suited
for convolutional architectures.
The measured data of each sensor only depends on the relative position to the robot.
In order to exploit this translational invariance,
a CNN is applied to predict the robot localization.

Similarly to the RF approach, the \sensorf data is normalized before utilization
in the \ac{CNN}. 
The values of each of the 10 sensor outputs are normalized 
to zero mean and unit variance by the following equation
\begin{equation}
    \vec{x}_i' = \frac{\vec{x}_i - \langle\vec{x}\rangle_{i}}{\vec{s}_{i}}
\end{equation}
where $\langle\vec{x}\rangle$ are the average sensor values and $\vec{s}$ are the
standard deviations computed over the training data.
This normalization facilitates the training of the neural network.
In addition, only the magnetometer and \rssi\, are used as these sensors
provide the most information for the localisation task.

A standard \ac{CNN} architecture with $3\times3$ convolution kernels and 
exponential linear units (elu)
as activation functions is used.
The \ac{CNN} is set up to estimate the $x$- and $y$-coordinates of the robot position
in addition to the uncertainty on these quantities. 
An asymmetric Gaussian likelihood:
\begin{equation}
    f(x|\mu, \sigma, r) =
    \begin{cases}
        N \cdot \exp{\left(-\frac{(x-\mu)^2}{2\sigma^2}\right)}, \qquad x \leq \mu \\
        N \cdot \exp{\left(-\frac{(x-\mu)^2}{2(\sigma r)^2}\right)}, \qquad \mathrm{otherwise} \\
    \end{cases}
\end{equation}
is used a loss function for each of the coordinates.
Therefore, the \ac{CNN} outputs a total of six values consisting of the estimated
position $(x, y)$ and its uncertainty, which is parameterized by $(\mu_x, \mu_y, r_x, r_y)$.
A custom activation function 
\begin{equation}
    \begin{split}
        \vec{\mu}' &= \vec{\mu} \\
        \vec{\sigma}' &= \exp{\left(\vec{\sigma}\right) + 0.001} \\
        \vec{r}' &= \exp{\left(\vec{r}\right) + 0.001} \\
    \end{split}
\end{equation}
is applied on the output of the neural network to enforce a positive lower bound on the uncertainties,
which helps to stabilize the training.
Further details on the network architecture are provided in table \ref{tab:CNN_architecture}.

\begin{table}[]
\centering
\caption{The layers of the \ac{CNN} architecture are illustrated. 
All convolution layers apply "SAME" padding and pooling layers 
utilize a $2\times2$-kernel.} 
\label{tab:CNN_architecture}
\begin{tabular}{lll}
\hline
\textbf{Layer Type} & \textbf{\# Units/Kernels} & \textbf{Activation} \\ \hline
2D Convolution      & 64                        & elu                 \\
2D Convolution      & 64                        & elu                 \\
2D Convolution      & 64                        & elu                 \\
2D Average Pooling  & -                         & -                   \\
2D Convolution      & 64                        & elu                 \\
2D Convolution      & 64                        & elu                 \\
2D Convolution      & 64                        & elu                 \\
2D Average Pooling  & -                         & -                   \\
2D Convolution      & 64                        & elu                 \\
2D Convolution      & 64                        & elu                 \\
2D Convolution      & 64                        & elu                 \\
2D Average Pooling  & -                         & -                   \\
Flatten             & -                         & -                   \\
Dense               & 128                       & elu                 \\
Dense               & 128                       & elu                 \\
Dense               & 6                         & custom             
\end{tabular}
\vspace{-0.5cm}
\end{table}

The \ac{CNN} is trained to predict the robot localisation for each frame individually based on the
data of each of the $23 \cdot 15$ sensors.
As a result, temporal information is not utilized.
However, the robot localization between consecutive frames is highly correlated due to constraining
physical laws of classical mechanics.
The robot is capable of a finite velocity and acceleration. 
Thus, its trajectory is continuous in time and governed by Newton's laws of motion.

The \ac{CNN} architecture as defined above is not capable of exploiting this information.
Instead of modifying the network architecture, a post-processing step is defined which penalizes
discontinuities between consecutive robot localisation predictions. The robot positions~$(\vec{x}, \vec{y})$ for $N$ consecutive frames are fit
by maximizing a regularized asymmetric Gaussian likelihood
\begin{equation}
    \begin{split}
        \mathcal{L}(\vec{x}, \vec{y} | \vec{\mu_x}, \vec{\mu_y}, \vec{\sigma_x}, \vec{\sigma_y}, \vec{r_x},\vec{r_y}) = \\
        \prod_i^N 
        \left( 
            f(\vec{x}_i | \vec{\mu_x}_i, \vec{\sigma_x}_i, \vec{r_x}_i) \cdot f(\vec{y}_i | \vec{\mu_y}_i, \vec{\sigma_y}_i, \vec{r_y}_i)
        \right)
        \cdot
        \mathrm{Reg}\left( \vec{x}, \vec{y} \right),
    \end{split}
\end{equation}
where $\mathrm{Reg}$ penalizes nonphysical acceleration and velocity values between consecutive frames.
The range of physical acceleration and velocity values are obtained from the label distribution over the training data.
Values exceeding beyond the boundaries of these distributions are nonphysical and must therefore be penalized.

In the context of this paper, $\mathrm{Reg}$ is chosen to penalize the absolute velocity~$\vec{|v|}$ and acceleration values~$\vec{|a|}$ via
\begin{equation}
    \lambda_v(|v|, c_v) = 
    \begin{cases}
        \exp{\left( 10 \cdot (|v| - c_v) \right)}, \qquad |v| > c_v \\
        0, \qquad \mathrm{otherwise} \\
    \end{cases}
\end{equation}
\begin{equation}
    \begin{split}
        \lambda_a(|a|, c_a) &=  - 2 c_a +
        \begin{cases}
            \exp{\left( 3 \cdot |a| \right)}, \qquad |a| > c_a \\
            \exp{\left( 2 c_a + |a| \right)}, \qquad \mathrm{otherwise} \\
        \end{cases} \\
    \end{split}
\end{equation}
respectively.
The acceleration and velocity values are computed via
\begin{equation}
    \begin{split}
        |v| = \frac{\sqrt{\Delta x^2 + \Delta y^2}}{\Delta t} \qquad |a| = \frac{|v|}{\Delta t} \\
    \end{split}
\end{equation}
and the values for $c_v$ and $c_a$ are chosen based on the time difference $\Delta t$ of $0.23 \si[]{\s}$ between
consecutive frames.

Future improvements to the neural network architecture may include 
a more adequate handling of timing information. 
Accurate accounting of timing information and the correlation between 
consecutive frames is crucial to achieving a precise robot localization.
The introduced regularization term attempts to include this information.
Alternative options such as recurrent neural networks are capable of
directly utilizing this information in contrast to the employed 2D \ac{CNN}. 



\subsection{Experimental Results}

The models are evaluated using the euclidean error metric and qualitatively assessed using the predicted trajectory.
Due to the limited space, the indoor localization is executed the euclidean error is in this case a well suited error metric for quantitative evaluation purposes. 
Criteria for qualitative evaluation include correctness and smoothness of the path compared to the ground truth.
Given the ground truth $\y$ and the predicted positions $\hat{\y}$ the euclidean distance between two vectors is defined as:
\begin{equation}
    d_{eucl}(\vec{y}, \hat{\vec{y}}) = \sqrt{\sum_{j=1}^d(y_j - \hat{y}_j)^2}.
\end{equation}
Using this distance the total error is defined as the average error across all predictions:
\begin{equation}
    \epsilon_{eucl} = \frac{1}{N} \sum_{i=1}^N d(\vec{y}^i, \hat{\vec{y}}^i).
\end{equation}

\begin{figure}
    \centering
    \includegraphics[width=\linewidth,height=4.55cm]{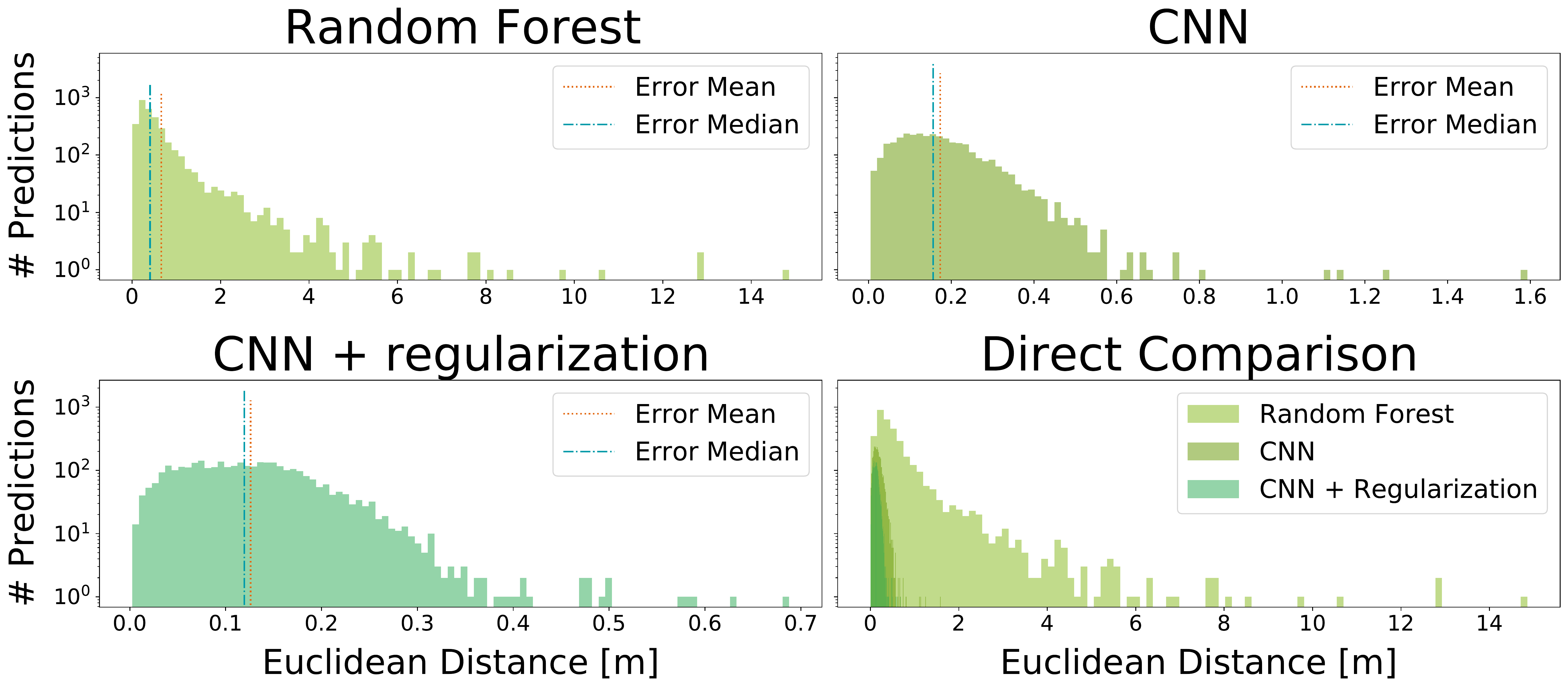}
    \caption{Error distribution histogram for the three trained models evaluated on the test data set. 
    The right bottom plot shows all three error distributions in direct comparison on a shared x-axis.
    Average and median error are shown as vertical lines. The y-Axis is shown in log-scale for improved outlier visualization. 
    Note that the x-axis scale is different for all three plots. The variance for each of the three models is: $\sigma^2_{RF}=0.85$, $\sigma^2_{CNN}=0.011$, $\sigma^2_{RCNN}=0.005$
    }
    \vspace{-0.5cm}
    \label{fig:error_dist}
\end{figure}


The error histograms for each model (note that each x-axis is scaled differently) are illustrated in \ref{fig:error_dist}. 
By using a logarithmic scale for the y-axis, outliers have improved visibility to better highlight the prediction variance.
Additionally, the bottom plot results from all three models on a share x-axis.
The error distributions clearly show the improvements from the traditional Random Forest model towards the \ac{CNN} and finally achieving the best result by introducing the additional preprocessing step as described in sub section \ref{ssec:CNN}.

\begin{figure}[htbp]
    \centering
    \vspace{-0.4cm}
    \includegraphics[width=\linewidth]{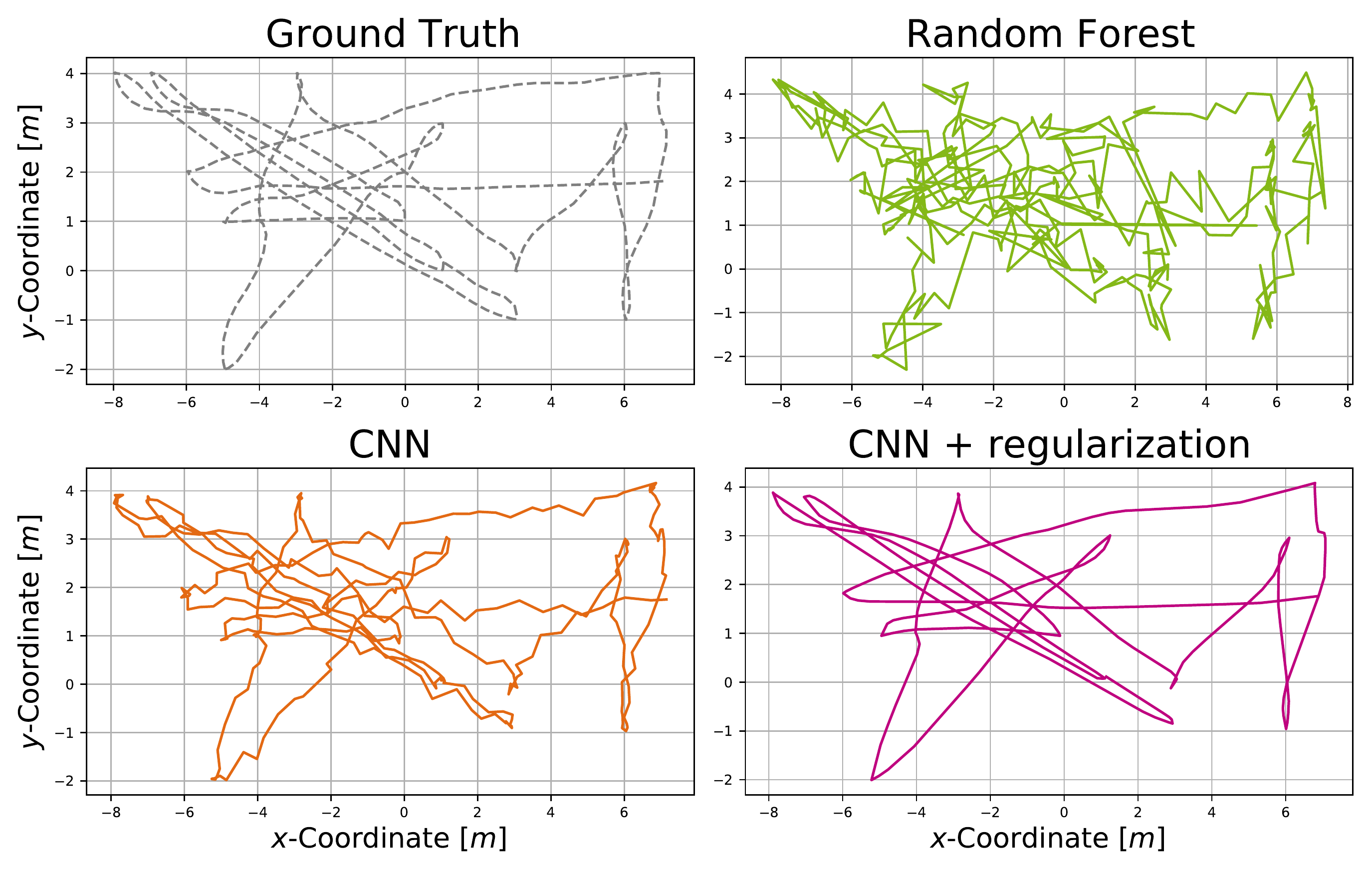}
    \caption{The true and predicted robot trajectories are shown for one run of the validation data set. 
    The regularization results in a smoother track.}
    \vspace{-0.4cm}
    \label{fig:regularized_trajectory}
\end{figure}

\paragraph{Random Forest}
While random forest models perform the worst out of the three models, they provide a solid baseline, showcasing that the presented data collection and preprocessing methods are empirically sound.
Predictions by the random forest are not constrained by real-world conditions such as acceleration and velocity. 
Therefore, the predicted trajectory shown in picture \ref{fig:regularized_trajectory} tends to occasionally exhibit erratic behaviour and zig-zagging, especially when some of the predictions are off the ground truth by as much as \SI{15}{\meter} in the worst case.

\paragraph{Convolutional Neural Network}
The \ac{CNN} improves upon the smoothness, decreasing the average euclidean distance, while also reducing the frequency of outliers and their magnitude.
Convolutions applied to the input data of $23\times15$ nodes appear to better extract spatio-temporal relationships in node-neighborhoods, leading to a more robust regressor with an improved average error and less severe outliers.
While the trajectory generated by the \ac{CNN} is visibly more smooth than that of the random forest it still exhibits some local zig-zagging.
However, the \ac{CNN} trajectory clearly resembles the ground truth trajectory.

\paragraph{Regularized CNN}
The effect of the regularization on the estimated robot trajectory is shown in the bottom right of figure \ref{fig:regularized_trajectory}.
Due to the additional penalization of nonphysical velocities and accelerations as well as the uncertainty quantification, outliers are greatly reduced and the trajectory is smoothed.
When compared to the ground truth, the regularized trajectory looks almost identical in terms of smoothness and positonal information.
The regularized \ac{CNN} outperforms the other models on average by  $\approx \SI{47}{cm}$ compared to the random forest and $\approx \SI{8}{cm}$ when compared to the vanilla \ac{CNN}.

Overall, the three models solve the localization task reasonably well with the regularized \ac{CNN} outperforming both the random forest and \ac{CNN}.
By applying two different machine learning models and incorporating additional physical real-world assumptions we have shown that our data collection and pre-processing pipeline is empirically sound.

\section{Conclusion} \label{conclussion}

In this work, we presented \emph{Sensor Floor}, a new hardware and software platform for distributed sensing. We used the platform to develop a robot localization application using machine learning with an objective to demonstrate diverse object identification in material handling facilities. The platform is arranged in asynchronous nodes with ground-truth data captured from a \vicon system. Additional pre-processing steps were employed to create synchronized data. Two models were trained using the generated data, one traditional random forest and one \acf{CNN}. A regularization post-processing step was added to the \ac{CNN} to create a smooth trajectory from the predicted positions. Future research is to develop diverse object identification with their vibration and RF footprint. By establishing an empirically sound, robust pipeline and providing a baseline for the localization task, the data and CNN architecture can be extended to different warehouse scenarios, such as coarse position estimation of multiple objects in the materials handling hall. Our future work includes adding the ability of self training of the nodes without \vicon data. As the replacement, we can treat the \sensorf node grid position as the anchor node.
\section*{Acknowledgment}

This work has received funding from the German Federal Ministry of Education and Research (BMBF) in the course of the 6GEM research hub under grant number 16KISK038 and the ML2R project under grant number 01|S18038A.
\bibliographystyle{IEEEtran}
\bibliography{sample-base}

\end{document}